\pdfoutput=1

\documentclass[11pt]{article}

\usepackage[preprint]{acl}

\usepackage{times}
\usepackage{latexsym}

\usepackage[T1]{fontenc}

\usepackage[utf8]{inputenc}

\usepackage{microtype}

\usepackage{inconsolata}

\usepackage{graphicx}
\usepackage{hyperref}       
\usepackage{url}            
\usepackage{booktabs}       
\usepackage{amsfonts}       
\usepackage{nicefrac}       
\usepackage{microtype}      
\usepackage{xcolor}         
\usepackage{graphicx}
\usepackage{subfigure}
\usepackage{amsmath}
\usepackage{mathtools}
\usepackage{amsthm}
\usepackage{diagbox}
\usepackage{multirow}
\usepackage{wrapfig}
\usepackage{algorithm}
\usepackage{algorithmic}
\usepackage[capitalize,noabbrev]{cleveref}
\usepackage{CJKutf8}

\usepackage{nicefrac}       
\usepackage{microtype}      
\usepackage{xcolor}         
\usepackage{comment}
\usepackage{mathtools}
\usepackage{subcaption}
\usepackage{tablefootnote}
\usepackage{textcomp} 
\usepackage{scalerel}
\usepackage{wrapfig}
\usepackage{colortbl}
\usepackage{longtable}
\usepackage{pifont}       
\usepackage{bbding}

\title{PRIV-QA: Privacy-Preserving Question Answering for Cloud Large Language Models}

\author{Guangwei Li\thanks{Contributed Equally.}, Yuansen Zhang\protect\footnotemark[1]\thanks{Work done during internship at Ant Group.} , Yinggui Wang \\ 
{\bf Shoumeng Yan, Lei Wang, Tao Wei}\\
  Ant Group \\
  \texttt{\{yunxuan.lgw,yuansen.zys\}@antgroup.com} }

\begin{document}
\maketitle
\begin{abstract}
The rapid development of large language models (LLMs) is redefining the landscape of human-computer interaction, and their integration into various user-service applications is becoming increasingly prevalent.
However, transmitting user data to cloud-based LLMs presents significant risks of data breaches and unauthorized access to personal identification information.
In this paper, we propose a privacy preservation pipeline for protecting privacy and sensitive information during interactions between users and LLMs in practical LLM usage scenarios. 
We construct SensitiveQA, the first privacy open-ended question-answering dataset. It comprises 57k interactions in Chinese and English, encompassing a diverse range of user-sensitive information within the conversations.
Our proposed solution employs a multi-stage strategy aimed at preemptively securing user information while simultaneously preserving the response quality of cloud-based LLMs.
Experimental validation underscores our method's efficacy in balancing privacy protection with maintaining robust interaction quality.
The code and dataset are available at \url{https://github.com/ligw1998/PRIV-QA}.

\end{abstract}

\section{Introduction}
\label{sec:introduction}

Large language models (LLMs) have marked a significant advancement in natural language understanding and generation, revolutionizing various industries, from customer service to content creation \cite{GPT-4, llama, llm_survey}. 
However, due to commercial demands and the models' enormous parameter sizes, many companies deploy cutting-edge LLMs on remote cloud infrastructures, offering services exclusively through APIs \cite{GPT-4, Claude}.
This requires users to transmit data to these cloud-based models, increasing the risks of personal sensitive information leaks \cite{Yao_2024, das2024securityprivacychallengeslarge}.
Therefore, ensuring the privacy and security of user data during the interactions is essential.

\begin{figure}[!t]
  \centering
  \includegraphics[width=\linewidth]{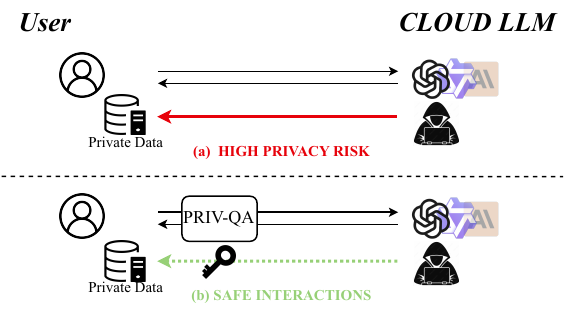}
  \caption{When using cloud-based LLMs, users must send queries to providers who can access their private data. However, by integrating our plug-and-play PRIV-QA framework, we can maximize privacy protection while ensure high-quality responses.
  }
  \label{fig:example}
\end{figure}


Previous privacy-preserving techniques for cloud API-based LLMs can be primarily divided into two categories.
Local Differential Privacy (LDP) methods ~\cite{SanText, CusText, InferDPT} sequentially substitute tokens in the text with new ones using specific mapping functions. 
However, these methods only protect user queries and do not safeguard the responses generated by the LLM.
On the other hand, text sanitization approaches \cite{kan2023protecting, shen2024fire, HAS} hide the private entities in the query for anonymization. However, they concentrate only on specific domains or tasks and limit their adaptability to general question-answering scenarios.
In addition, the none-privacy entities (\textit{for example, the world's richest man}) may also leak private information.

In this paper, we introduce \textbf{PRIV-QA} to safeguard user \textbf{PRIV}acy during \textbf{Q}uestion \textbf{A}nswering when interacting with cloud LLMs.
First, we construct SensitiveQA, the first bilingual general privacy question-answering dataset, containing over 57k interactions between users and cloud LLMs in both Chinese and English.
Each query in SensitiveQA includes a background text rich in personal sensitive information, followed by a final question.
Second, we propose a privacy-preserving framework to prevent malicious attacks during interactions with cloud-based LLMs.
Specifically, our framework incorporates a multi-stage text sanitization pipeline that classifies each word or token in a query into three distinct privacy and importance levels and subsequently applies tailored protection mechanisms to each term based on its assigned level before being transmitted to cloud-based LLMs.
After obtaining the responses from cloud LLMs, the \textit{recover module} ensures the restoration of sensitive information, corrects potential reasoning or conclusion errors, and delivers the final recovered response to the user.

Experiments on the SensitiveQA dataset demonstrate that \textbf{PRIV-QA} achieves recall rates of $89.40\%$ and $73.01\%$ in sensitive information detection for English and Chinese, respectively.
Additionally, \textbf{PRIV-QA} can resist $85.83\%$ of extraction attacks while achieving the highest scores (for example, a BLEU score of $0.563$ in English) for the recovery responses.

Our contributions are summarised as follows:
\vspace{-0.2cm}
\begin{itemize}
    \item We construct SensitiveQA, the first general privacy question-answering dataset, containing 57k interactions between users and cloud LLMs in either Chinese or English.
\vspace{-0.2cm}
    \item We propose a privacy-preserving framework comprising a multi-stage text sanitization procedure to eliminate and restore sensitive information during interactions.
\vspace{-0.2cm}
    \item Experimental results demonstrate that \textbf{PRIV-QA} outperforms all baselines in detecting sensitive information and query protection while also maintaining the highest utility of user queries.
\end{itemize}

\section{Related Work}
\label{sec:related work}
\subsection{Privacy Preserving Techniques}
Privacy concerns have become a significant issue for the broad adoption of LLMs, and a variety of works have explored the privacy-preserving techniques for LLMs \cite{yan2024protectingdataprivacylarge, edemacu2024privacypreservingpromptengineering}.
Differential Privacy (DP) \cite{InferDPT, utpala-etal-2023-locally, tang2024privacypreservingincontextlearningdifferentially, duan2023flocksstochasticparrotsdifferentially} is widely adopted by adding random noise into the dataset.
Federated Learning \cite{chen2023federatedlargelanguagemodel, yu2024federatedfoundationmodelsprivacypreserving, zhang2024buildingfederatedgptfederated} offers a local-cloud collaboration paradigm without the need to send personal data to the central server.
Other approaches like Homomorphic Encryption \cite{chen2022thexprivacypreservingtransformerinference, hao2022iron} and Multi-Party Computation (MPC) \cite{goldreich1998secure, dong2023pumasecureinferencellama7b} is time-consuming and can hardly be applied in real-world scenarios.

\subsection{Text Sanitization Approaches}

For cloud models with only API provided, the above approaches can be infeasible to implement.
Therefore, text sanitization techniques rise as an effective method, which aims to identify and eliminate sensitive information from the text.
\cite{CusText, SanText, InferDPT} replace tokens selective in the text, however, their methods are not focused on privacy attributes and still lead to privacy leakage risks.
\cite{HAS} hide the private entities for anonymization and seek private entities for de-anonymization. However, they only focus on classification and translation tasks, which can not be generalized to other tasks.
\cite{lin2024emojicryptpromptencryptionsecure} uses Emoji to encrypt the user
inputs before sending them to LLM.
\cite{kan2023protecting} utilizes a privacy filter module to identify and replace the privacy information in the text without obfuscating non-privacy entities. In addition, they use the open-sourced model in the filter module and can fail for long documents without training.
On the contrary, our method is a multi-stage privacy-protecting framework for general QA scenarios and our hide and recover modules have been trained on corresponding tasks to better align with the requirements.

\section{SensitiveQA Dataset}
\label{sec:dataset}

In order to simulate real-world user interactions with cloud-based LLMs, we construct a dataset that encompasses a variety of dialogues that contain personal privacy information.
Typically, an ordinary user query can be divided into two components: background text and final question.
The background text can vary in length and form.
It may consist of previous chat dialogue, passages retrieved from a local knowledge base, or complex user-modified instructions.
Final questions are generally connected to this background information. 
It is important to note that both the background text and the questions may contain personally sensitive information.
Directly sending this information to cloud LLMs may cause privacy issues.

In this paper, we focus on the general question-answering task in both Chinese and English.
For Chinese, we collect news and wiki terms in News Summarization\footnote{https://www.kaggle.com/datasets/sbhatti/news-summarization}, CLTS~\cite{liu2022clts+}, WikipediaCN~\cite{wikidump}, and randomly select 10200 background texts.
For English, we collect 4434 background texts from a personal identification information dataset\footnote{https://www.kaggle.com/datasets/alejopaullier/pii-external-dataset}.
For each background text, we use OpenAI GPT-4o \cite{GPT-4} to generate a range of questions, encompassing various tasks such as information extraction, open-ended Q\&A, and text summarization.
More details can be found in Appendix \ref{appendix: SensitiveQA Dataset Construction}.

~\Cref{tab:dataset} presents a comparison between SensitiveQA and currently available public datasets.
General instruction-tuning datasets like Alpaca~\cite{alpaca} lack sufficient user-related privacy information.
In addition, news datasets such as CNN/Daily Mail~\cite{see-etal-2017-get} and named entity recognition (NER) datasets like MSRA~\cite{levow-2006-third} contain background texts but do not have related questions to form a comprehensive LLM query.
On the other hand, privacy-focused datasets employed in prior research \cite{HAS} are tailored to specific domains or tasks, thereby limiting their adaptability to open-ended question-answering scenarios.
Therefore, SensitiveQA offers a dataset that is both privacy-related and general enough to be applied across a wide range of QA contexts.
\definecolor{baselinecolor}{gray}{.92}

\newcommand\cmark {\textcolor{green}{\ding{52}}}
\newcommand\xmark {\textcolor{red}{\ding{55}}}

\begin{table}[tbp]

  \centering
   \begin{resizebox}{\columnwidth}{!}
   {
    \begin{tabular}{ cc c c c}
      \toprule
      \textbf{Dataset} & \textbf{Num} & \textbf{Privacy} & \textbf{QA} & \textbf{Lang} \\
      \midrule
        Alpaca   &  $52000$  & \xmark & \cmark &  EN \\
        MSRA &  $50729$  & \xmark  & \xmark&  CN \\
        CNN/Daily Mail & $311672$ & \cmark & \xmark &  EN  \\
        MedQA  &  $46974$  & \xmark & \cmark &  CN\&EN \\  
        HaS Synthetic  & $19703$  & \cmark  & \xmark &  CN\&EN \\    
         \rowcolor{baselinecolor}SensitiveQA  &  $57251$  &   \cmark  & \cmark &  CN\&EN  \\
      \bottomrule
    \end{tabular}%
   }
   \end{resizebox}
     \caption{A comparison between the SensitiveQA Dataset with other public datasets.}
  \label{tab:dataset}
\end{table}

By leveraging the SensitiveQA dataset, our framework is trained to effectively safeguard user queries while preserving the high-quality response capabilities of LLMs. We conduct a comprehensive evaluation of our proposed method in open-ended question-answering scenarios.

\begin{figure*}[tbp]
  \centering
  \includegraphics[width=\linewidth]{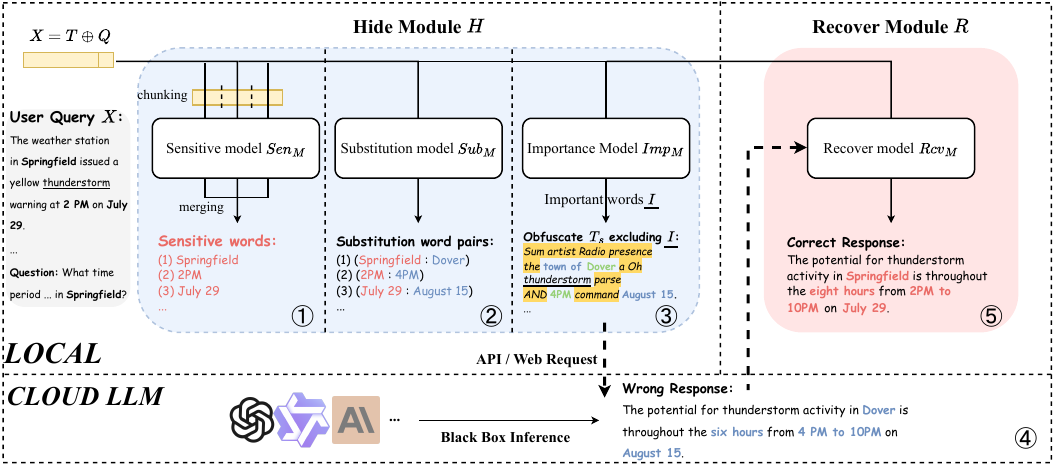}
  \caption{Workflow of our privacy-preserving framework. For a user query, the hide module performs sensitive information detection and substitution. Then, we obfuscate the non-privacy text excluding certain important words. The recover module obtains the LLM responses and restores the privacy information in the text.}
  \label{fig:framework}
  \vspace{-0.2cm}
\end{figure*}

\section{Method}
\label{sec:method}

Most frontier LLMs, which companies and individual users widely use, are close-sourced, including ChatGPT~\cite{GPT-4}, Qwen-Series~\cite{qwen2}, and Claude~\cite{Claude}.
These LLMs are typically hosted on remote cloud computing infrastructures and are exclusively accessible through web interfaces or APIs, rendering them a black box to users.
To mitigate privacy risks in cloud-based LLM applications, a viable solution is to de-identify the users' queries before sending them to cloud LLMs and post-processing the LLM's responses to restore the sensitive information.

Our proposed \textbf{PRIV-QA} framework comprises two core components: (1) Hide Module $H$ designed for de-identifying private information, and (2) Recover Module $R$ responsible for restoring the original sensitive data. The overview and mechanism of our framework are presented in ~\Cref{fig:framework} and \Cref{alg:hide}.
This section elaborates on the framework's constituent elements and elucidates their respective functionalities.

\subsection{Hide Module: Query Protection by Privacy Level}
\label{subsec:hide}

\begin{figure}[!t]
  \centering
  \includegraphics[width=\linewidth]{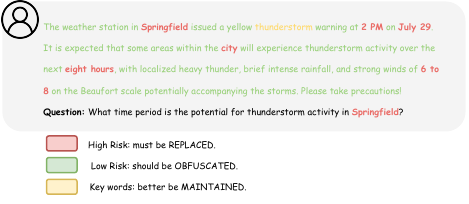}
  \caption{An example of tri-level classification to a given user query.
  }
  \label{fig:level}
    \vspace{-0.2cm}
\end{figure}

We implement a multi-stage text sanitization pipeline to protect sensitive information in user queries.
For a user query $X$, our hide module dynamically implements tailored protection strategies based on the identified risk levels of individual tokens.
Specifically, we establish a tri-level classification (High-Risk, Low-Risk, and Key-Words in~\Cref{fig:level}) that evaluates each lexical component based on its potential privacy implications and semantic importance.
Distinct protection mechanisms are then applied according to these classifications.
Following we provide a detailed exposition of our pipeline and implementation.

\paragraph{Sensitive Information Detection}

For a general user query $X = T \oplus Q$, compromised of background text $T$ and a question $Q$, we define words that contain personal privacy information as High-Risk words $S$.
Through reviewing the General Data Protection Regulation (GDPR\footnote{https://eur-lex.europa.eu/eli/reg/2016/679/oj/eng}), we identify five categories of privacy information: (1) Name of an individual or group; (2) Dates and times; (3) Locations and specific places; (4) Personal information including phone numbers, id, (email) addresses, titles, etc; (5) Sensitive numbers (monetary values, measurements, percentages, etc).

We finetune a generative model $Sen_M$ to detect the sensitive words in $X$.
Detailed prompts can be found in Appendix \ref{prompt: sensitive words detection}.
To construct the training data, we prompt GPT-4o \cite{GPT-4} to detect sensitive words for a given query. We manually verify the generations and obtain 5000 samples for training.

However, directly processing the entire query $X$ with $Sen_M$ can be challenging and lead to the omission of sensitive words, particularly when the background text $T$ is extensive. To address this issue, we segment $X$ into smaller chunks, $x_i$, each constrained by a token limit. Each chunk $x_i$ is then processed by $Sen_M$ to produce a list of sensitive words, $s_i$. Finally, we aggregate and deduplicate the sensitive words identified across all chunks to form a comprehensive set of sensitive words $S$.


\paragraph{Sensitive Words Substitution}
To protect the privacy information in $X$, each word in the sensitive word set $S$ must be replaced with a semantically similarity but entirely different word to ensure user privacy. However, simply substituting sensitive words with random words or meaningless placeholders can severely disrupt the sentence's semantics and structure, leading to incoherent or irrelevant responses from the cloud LLMs. 
To address this, we utilize another model $Sub_M$ to generate a privacy-preserving substitute $p_i$ for each sensitive word $s_i$. 
$Sub_M$ ensures that the substituted query remains semantically coherent and structurally intact. The word pairs $(s_i : p_i)$ are then used to transform the original query $X$ into a privacy-protected version $X_s = T_s \oplus Q_s$, where all user-sensitive information is effectively concealed. 

The training data for $Sub_M$ is generated by prompting GPT-4 \cite{GPT-4}, resulting in a dataset of 55,000 samples. Further details on the prompts used for this process can be found in Appendix \ref{prompt: sensitive words substitution}.


\paragraph{None-Privacy Text Obfuscation}

To further protect the low-risk part in $X$, we follow \cite{InferDPT} to substitute partial tokens in this part.
Specifically, for a certain LLM with the corresponding token set V. We first select a subset of $V$, defined as $V_{sub}$ with a size $K$. In addition, we have a differential privacy mechanism scoring function $u$ and randomized mechanism $M$,
For each token $t_i$ in $V_{sub}$, where $i = 1, 2, \dots, k$, we compute a random adjacency $C_W(t_i)$ based on the utility and privacy principles. 
Specifically, we obtain corresponding embedding $emb_i$ for $t_i$ and add noise following the Laplace mechanism and obtain $\hat{emb}_i$.
Then for each token in $V$ and with its embedding $emb$, we add it to $C_W(t_i)$ if it satisfies
\begin{equation}
    d(emb, emb_i) \leq d(emb_i, \hat{emb_i})
\end{equation}
where $d$ is a distance function.
For inference, we randomly select a token from $C_W(t_i)$ and replace $t_i$ in the text.

\paragraph{Impotrant Words Preservation}
Certain words play a crucial role in understanding the text and generating an appropriate response to the question. For example, in medical diagnosis scenarios, obscuring privacy-sensitive terms such as disease could lead to inaccurate diagnostic outcomes. To address this, we employ a model $Imp_M$ to identify and preserve such critical terms, ensuring they remain unobfuscated. 

\begin{algorithm}[tbp]
    \caption{PRIV-QA Framework}
    \label{alg:hide}
    \renewcommand{\algorithmicrequire}{\textbf{Input:}}
    \renewcommand{\algorithmicensure}{\textbf{Output:}}
    \begin{algorithmic}[1]
        \REQUIRE Background Text $T$, Question $Q$, Full Query $X = T \oplus Q$, Sensitivity Model $Sen_M$, Substitution Model $Sub_M$, Importance Model $Imp_M$, Recover Model $Rcv_M$;
        \ENSURE Correct Response $A$;    
        \STATE Split $X$ into $n$ chunks $X=x_0\oplus x_1...\oplus x_n$;
        \FOR {each $x_i$}
            \STATE Sensitive words $s_i = Sen_M(x_i)$;
        \ENDFOR
        \STATE $S = s_0 \cup s_1 ... \cup s_n$;
        \STATE Substitution Word Pairs $P=Sub_M(S)$;
        \STATE Use word pairs $P$ to replace $X$ into $X_s = T_s \oplus Q_s$;
        \STATE Select important words $I=Imp_M(T_s,Q_s)$;
        \STATE (Optional) Obfuscate all tokens in $T_s$ into $T_{s,o}$ excluding $I, P$;
        \STATE $X' = T_{s,o} \oplus Q_s$;
        \STATE Get LLM Responses $A' = LLM(X') $;
        \STATE Get Correct Responses $A = R(X', X, A')$;
        \RETURN $A$
    \end{algorithmic}
\end{algorithm}


\subsection{Recover Module: Cloud Response Correction}
\label{subsec:recover}

The sanitized query $X'$ is securely transmitted to cloud-based LLMs, yielding a response $R'$.
However, due to the substitution and obfuscation applied to the query, some information in $R'$ is incorrect, resulting in a wrong reasoning chain and erroneous response.
To address this, we employ a generative LLM $Rcv_M$ to restore the substituted words $p_i$ back to the original words $s_i$ and correct the reasoning errors in $R'$ while keeping the overall thought process intact.
Afterward, we obtain the final correct response $R$.
More details for training $Rcv_M$ can be found in Appendix \ref{prompt: recover model}.



\section{Experiments}
\label{sec:experiments}

\subsection{Experiment Setup}
\label{subsec:setup}

\paragraph{Evaluation Setup and Test Set.} 
The proposed \textbf{PRIV-QA} is a bilingual framework supporting both Chinese and English, enabling users to interact with multiple cloud LLMs securely. We conduct evaluations for these two languages independently.

We employ the SensitiveQA dataset for both training and evaluation purposes. To comprehensively assess performance, we further partition the test set into two distinct tasks: (1) information extraction and (2) open-ended question answering. The first task is designed to evaluate the model's capability in identifying sensitive information, while the second task measures the impact of our \textbf{PRIV-QA} framework on the response quality of cloud-based large language models (LLMs).

\paragraph{Metrics.}  We adopt both metrics-based and model-based evaluation to judge the quality of responses from the LLM.
We use BLEU~\cite{papineni2002bleu}, METEOR~\cite{banerjee-lavie-2005-meteor}, and ROUGE~\cite{lin-2004-rouge} metrics to objectively measure the discrepancy between the recovered responses and the ground-truth answers, allowing us to evaluate the quality of the final recovered responses.
We define the ground-truth answer as the LLM response without any privacy protection.
For each test sample, we offer three ground-truth answers.
In addition, since Large Langauge models can perform human-like judgment \cite{zheng2023judgingllmasajudgemtbenchchatbot}, we perform model-based evaluations to evaluate the generation quality comprehensively. We choose OpenAI GPT-4o as a subjective evaluator to determine which response better addresses the user query and whether the information provided is accurate and aligned with the original query.
For each (recovered response, ground-truth) pair, GPT-4o will give a win/tie/lose judgment.
To ensure reliability, we manually verify the evaluation results and confirm that they align with human judgment in over $90\%$ of the cases.
Detailed prompts can be found in ~\Cref{Appendix: GPT-4 Judge Prompt}.

\paragraph{Implementation.} We choose the open-source Qwen2-Chat ~\cite{qwen2} model as our base model because of its relatively lightweight and strong performance in both Chinese and English. 
We choose Qwen2-0.5B-Chat for $Sen_M$, $Sub_M$ and $Imp_M$, and Qwen2-1.5B-Chat for $Rec_M$.
In addition, for cloud-based LLM, we select GPT-4-turbo \footnote{https://platform.openai.com/docs/models/o1} and Qwen-Plus \footnote{https://bailian.console.aliyun.com/\#/home}.
For token adjacency generation, we follow \cite{InferDPT} to generate OpenAI GPT-4-turbo token adjacency utilizing text-embedding-ada-002 \footnote{https://openai.com/index/new-and-improved-embedding-model/} APIs. For Qwen-Plus, we utilize Qwen2-7B-Chat for token adjacency generation.
More details can be found in ~\Cref{appendix: Experimental Details}.

\subsection{Results}
\label{subsec:results}

\begin{table}[tbp]
  \centering
   \begin{resizebox}{1.0\columnwidth}{!}
   {%
\begin{tabular}{l cc cc cc}
\toprule
 & \multicolumn{4}{c}{SensitiveQA} \\
\cmidrule(l){2-5}  
  & \multicolumn{2}{c}{EN} &\multicolumn{2}{c}{CN}\\
  & Precision &Recall &Precision &Recall\\
\midrule
DeBERT  & $74.13$ & $72.70$  & $37.09$ & $19.59$ \\
HaS  & $61.78$ & $39.88$ & $50.28$ & $60.63$\\
\midrule
\rowcolor{baselinecolor}\textbf{PRIV-QA} $Sen_M$  & $\textbf{91.36}$& $\textbf{89.40}$ & $\textbf{78.99}$ & $\textbf{73.01}$\\
\bottomrule
\end{tabular}%
   }
   \end{resizebox}
     \caption{Results for sensitive information detection. We provide precision and recall for three methods. The best results are \textbf{bolded}.}
  \label{tab:sensitiveresults}
\end{table}

In order to do a comprehensive observation of the framework performance in each stage, we evaluate our proposed method from three aspects: (1) Sensitive information detection performance; (2) The overall level of protection for the processed query transmitted to the cloud; (3) The utility, correctness, and quality of the response.

\paragraph{Sensitive Information Detection Performance.} 
Correctly identifying the sensitive and privacy-related information in the user query is a prerequisite for successfully protecting the query. 
We compare our sensitive detection model $Sen_M$ with Hide-and-Seek (HaS) ~\cite{HAS}, which uses a combination of spaCy \footnote{https://github.com/explosion/spaCy} and ltp \footnote{https://github.com/HIT-SCIR/ltp}.
We also utilize another finetuned DeBERT-based model as a fundamental baseline.

We construct a test set for evaluation following the same prompts in ~\Cref{prompt: sensitive words detection}. 
As shown in \Cref{tab:sensitiveresults}, our sensitive detection model achieves superior performance compared to previous methods in both precision and recall metrics. This indicates its enhanced capability to accurately identify sensitive words within queries in complex real-world scenarios.

\begin{table}[tbp]
  \centering
   \begin{resizebox}{\columnwidth}{!}
   {%
\begin{tabular}{l cc cc cc}
\toprule
 & \multicolumn{4}{c}{SensitiveQA} \\
\cmidrule(l){2-5}  
  & \multicolumn{2}{c}{EN} &\multicolumn{2}{c}{CN}\\
  & EDR(\%)$\uparrow$ &BLEU$\downarrow$ &EDR(\%)$\uparrow$ &BLEU$\downarrow$\\
\midrule
CUSTEXT+  & $29.71$ & $0.1955$ & $53.30$ & $0.1617$  \\
SANTEXT+ & $37.67$ & $0.3199$ & $23.79$ & $\textbf{0.0269}$\\
HaS & $73.84$ & $0.9310$ & $75.54$ & $0.7468$\\
\midrule
\textbf{PRIV-QA} w/o $Obf$ & $81.74$ & $0.9336$ & $94.75$ & $0.7753$ \\
\rowcolor{baselinecolor}\textbf{PRIV-QA} w/ $Obf$ & $\textbf{85.83}$ & $\textbf{0.0371}$ & $\textbf{95.24}$ & $0.5632$ \\
\bottomrule
\end{tabular}%
   }
   \end{resizebox}
     \caption{Results for query protection. We report EDR (Extraction Defense Rate) and BLUE for each method. w/o $Obf$ refers to results without our non-privacy text obfuscation method. The best results are \textbf{bolded}.}
  \label{tab:hideresults}
\end{table}

\begin{table*}[!ht]
    \centering
   \begin{resizebox}{2.0\columnwidth}{!}
   {%
\begin{tabular}{lcccccccccc}
      \toprule

 & \multicolumn{5}{c}{English}                 & \multicolumn{5}{c}{Chinese}                   \\
     & \multicolumn{1}{c}{BLEU} & \multicolumn{1}{c}{METEOR} & \multicolumn{1}{c}{R-1} & \multicolumn{1}{c}{R-2} & \multicolumn{1}{c}{R-L} & \multicolumn{1}{c}{BLEU} & \multicolumn{1}{c}{METEOR} & \multicolumn{1}{c}{R-1} & \multicolumn{1}{c}{R-2} & \multicolumn{1}{c}{R-L} \\
   \cmidrule(lr){2-6} \cmidrule(lr){7-11}
       & \multicolumn{10}{c}{\textbf{GPT-4-turbo}}                  \\
\midrule
   
CUSTEXT+ & $0.377$ &  $0.622$  &  $0.644$   & $0.459$  &  $0.624$   & $0.204$ &  $0.325$ &  $0.391$   &  $0.224$   &   $0.342$  \\
SANTEXT+  & $0.397$ &  $0.595$  &  $0.610$   & $0.422$    &  $0.592$   & $\underline{0.233}$ & $\underline{0.364}$   &   $\underline{0.447}$  & $\underline{0.273}$    &   $\underline{0.389}$  \\
HaS       &  $0.241$ &  $0.407$  &  $0.352$   &   $0.256$  &  $0.340$   & $0.126$ &  $0.283$  & $0.286$    &   $0.158$  &    $0.227$ \\
 \midrule
\textbf{PRIV-QA} w/o $Obf$   & $\textbf{0.641}$ & $\textbf{0.820}$   &  $\textbf{0.805}$   &  $\textbf{0.678}$   & $\textbf{0.792}$    & $\textbf{0.707}$ & $\textbf{0.792}$   &   $\textbf{0.796}$  &   $\textbf{0.662}$  &  $\textbf{0.734}$   \\
\rowcolor{baselinecolor}\textbf{PRIV-QA} w/ $Obf$  & $\underline{0.563}$ & $\underline{0.766}$  &  $\underline{0.762}$  &  $\underline{0.623}$   &  $\underline{0.747}$   & - &  -  & -  & -   & -   \\
   \midrule
       & \multicolumn{10}{c}{\textbf{Qwen-Plus}}                  \\
       \midrule
CUSTEXT+ &  $0.271$ &   $0.540$ &  $0.521$  & $0.306$  &   $0.483$  & $0.232$ &  $0.402$  &  $0.442$  & $0.227$  &   $0.346$ \\
SANTEXT+  & $0.466$ &  $0.689$  &  $0.669$  &  $0.493$   &  $0.637$   & $0.340$ &  $0.511$  &  $0.553$   &   $0.323$  &    $0.442$ \\
HaS       & $0.460$ &  $0.641$  &  $0.590$   &   $0.470$  &  $0.570$   & $0.105$ &  $0.253$  &  $0.266$   &   $0.122$  &  $0.193$   \\
 \midrule
\textbf{PRIV-QA} w/o $Obf$     & $\textbf{0.603}$ &  $\textbf{0.798}$  &  $\textbf{0.792}$   &  $\textbf{0.673}$   &  $\textbf{0.779}$   & $\textbf{0.596}$ &   $\textbf{0.727}$ &  $\textbf{0.722 }$  &  $\textbf{0.579}$   &  $\textbf{0.651}$   \\
\rowcolor{baselinecolor}\textbf{PRIV-QA} w/ $Obf$     & $\underline{0.521}$ &  $\underline{0.744}$  &  $\underline{0.733}$   &   $\underline{0.599}$  & $\underline{0.720}$    & $\underline{0.510}$  &  $\underline{ 0.674}$ &  $\underline{0.677}$   &    $\underline{0.515}$ &  $\underline{0.599}$   \\
        \bottomrule
\end{tabular}
   }
   \end{resizebox}

     \caption{Results for the recovery quality. We report BLUE, METEOR, ROUGE-1, ROUGE-2 and ROUGE-L for each baseline method. We select GPT-4-turbo and Qwen-plus as our cloud models. For GPT-4-turbo, the Chinese result for \textbf{PRIV-QA} w/ $Obf$ is not reported due to the lack of open-source support for Chinese tokens in \cite{InferDPT}. The best results are \textbf{bolded}, and the second best ones are \underline{}{underlined} (for prompting method).}
  \label{tab:recoverresults}
\end{table*}

\paragraph{The Query Protection Performance.} We evaluate query protection performance using two metrics: (1) Extraction Defense Rate (EDR) and (2) BLEU score between the protected query and the original input query. As outlined in ~\Cref{subsec:setup}, our test set comprises two tasks. Here, we leverage the information extraction task to simulate potential attacks from cloud LLM providers. For each sample, both the original query and its protected counterpart are paired with the same sensitive word extraction question. The cloud LLM generates respective sets of extracted results based on the extraction queries. We assess whether the two extracted results are consistent. If the cloud LLM extracts inconsistent words from the protected query, the protection is deemed successful. The Extraction Defense Rate (EDR) is calculated as the average defense success rate across all test queries, with a higher EDR indicating superior query protection performance. Additionally, we use the BLEU score as a supplementary metric to evaluate structural and semantic similarity. 

We compare the query protection performance of our \textbf{PRIV-QA} framework with local text sanitization methods, including CUSTEXT+~\cite{CusText}, SANTEXT+~\cite{SanText}, and the word substitution approach from HaS \cite{HAS}. As demonstrated in Table \ref{tab:hideresults}, \textbf{PRIV-QA} achieves the highest EDR in both English and Chinese, underscoring its superior query protection capabilities. Furthermore, our non-privacy text obfuscation process provides enhanced protection by making the query unreadable to humans, as evidenced by the lower BLEU scores. 
However, the higher protection does not lead to performance degradation compared with baseline methods as evidenced in Table \ref{tab:recoverresults}.
In contrast, CUSTEXT+ and SANTEXT+ exhibit lower EDR and higher BLEU scores, indicating their ineffectiveness in safeguarding sensitive entities.

\begin{figure*}[htbp]
  \centering
  \includegraphics[width=0.85\linewidth]{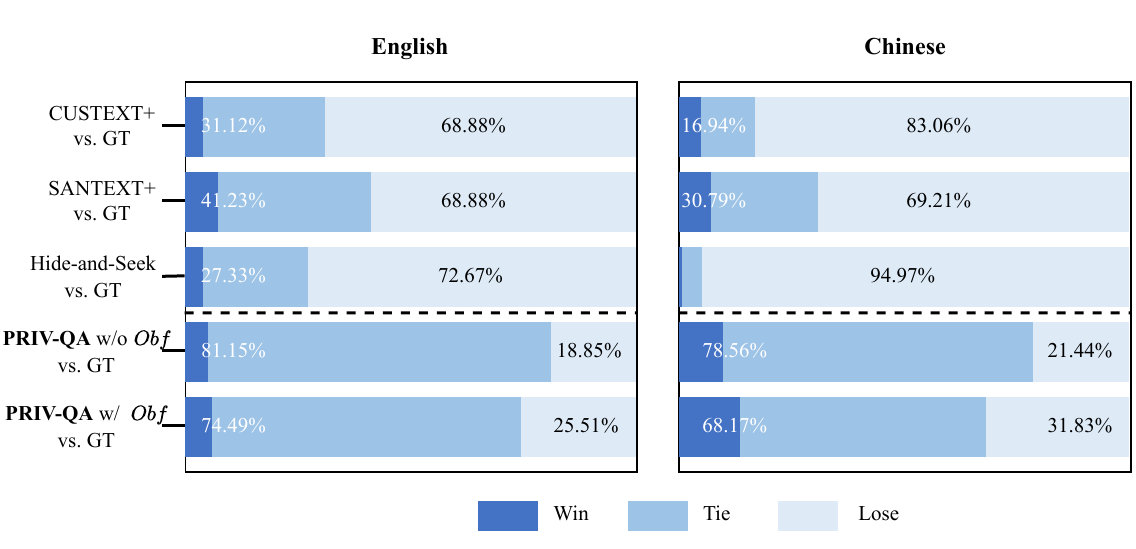}
  \caption{Final response quality results for each method compared to ground truth. GPT-4o is selected as the win-tie-lose judge. We report the \textbf{win\&tie} rate and \textbf{lose} rate. 
  }
  \label{fig:winrate}
      \vspace{-0.2cm}
\end{figure*}

\begin{figure}[htbp]
  \centering
  \includegraphics[width=0.95\linewidth]{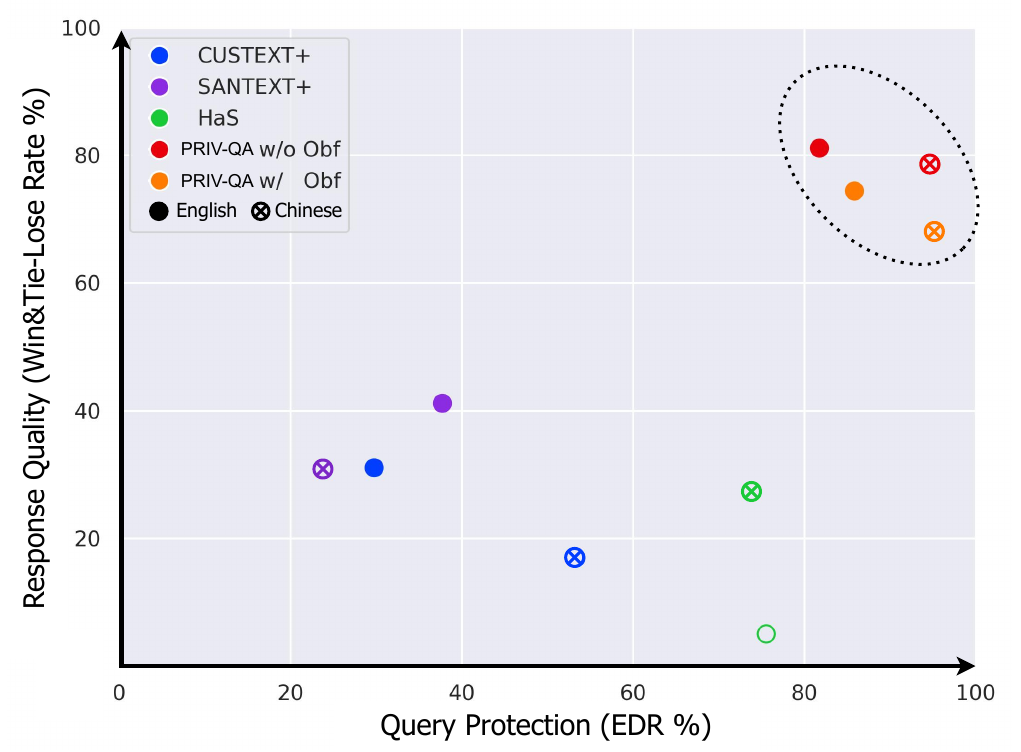}
  \caption{Trade-off between query protection and final response quality. The x-axis refers to the EDR (Extraction Defense Rate), and the y-axis refers to the win\&tie rate against ground truth judged by GPT-4o. Top-right methods perform best.
  }
  \label{fig:balance}
        \vspace{-0.2cm}
\end{figure}

\paragraph{Recovery Quality.} 
We assess the quality of the final recovered responses using both metrics-based and model-based evaluation methods, as detailed in Section \ref{subsec:setup}. Our approach is compared against CUSTEXT+ \cite{CusText}, SANTEXT+ \cite{SanText}, and HaS \cite{HAS}. For CUSTEXT+ and SANTEXT+, the responses are directly obtained from the cloud LLM, whereas for HaS, the responses are generated by its seek model, adhering to the hyper-parameters specified in its original paper.
We do not compare with \cite{kan2023protecting} as the methodology proposed in their work closely resembles HAS, and neither their model nor code has been made publicly available.

The metrics-based results are shown in ~\Cref{tab:recoverresults}. 
We can figure out that \textbf{PRIV-QA} outperforms previous methods on all metrics both in English and Chinese.
For instance, \textbf{PRIV-QA} achieves BLUE scores of 0.641 and 0.603 for GPT-4-turbo and Qwen-plus, respectively, surpassing HaS by 0.4 and 0.143 points.
This could be because HaS is task-specific (only for summarization and translation tasks) and, therefore, struggles in the general QA scenarios.
CUSTEXT+ and SANTEXT+ perform better than HaS but still lag significantly behind \textbf{PRIV-QA}.
Additionally, our non-privacy text obfuscation methods provide a stronger defense against extraction attacks (\Cref{tab:hideresults}), but result in a decline in slight response quality, such as a drop of 0.078 BLUE points for GPT-4-turbo in English.

The model-based evaluation by GPT-4o is presented in ~\Cref{fig:winrate}.
We report the sum of win and tie rates as positive rates.
According to ~\Cref{fig:winrate}, \textbf{PRIV-QA} with text obfuscation achieves a 74.49\% positive rate in English and 68.17\% in Chinese, greatly surpassing previous methods and indicating higher response quality.

\paragraph{Trade-off between Security and Quality.} Generally speaking, a higher security leads to a lower final response quality since the processed query is more challenging to understand, and the recovery process is also more difficult.
Under open-ended question-answering scenarios, we further investigate the security and the final response quality trade-off of each method in ~\Cref{fig:balance}.
With none-privacy text obfuscation, \textbf{PRIV-QA} achieves higher query protection with a minor sacrifice in recovered response quality.
Notably, all configurations of \textbf{PRIV-QA} demonstrate a substantial overall improvement over prior methods.

\paragraph{The Time Consumption of PRIV-QA}

To evaluate the efficiency of \textbf{PRIV-QA}, we conducted a comparative analysis of time consumption with and without the framework under varying input and output token numbers. 

\begin{figure}[tbp]
  \centering
  \includegraphics[width=0.95\linewidth]{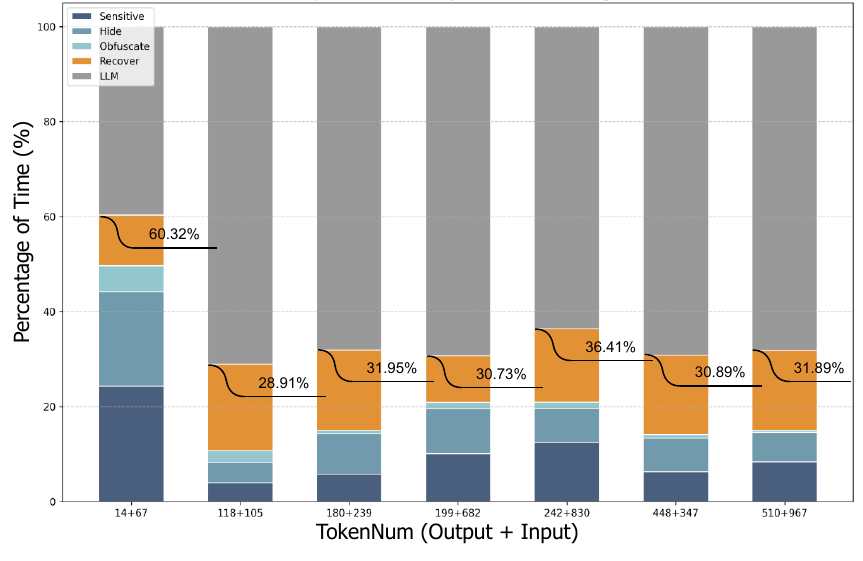}
  \caption{Proportion of time consumption per module in PRIV-QA (deployed locally on one A800 GPU) across different Input + Output token lengths.
  }
  \label{fig:time}
  \vspace{-0.2cm}
\end{figure}

As depicted in Figure \ref{fig:time}, when the number of input and output tokens is relatively small, the PRIV-QA framework introduces a time consumption increase of approximately 60\%. 
However, as the token count increases, the relative time overhead introduced by PRIV-QA diminishes, stabilizing around 30\%. 
This reduction occurs because the fixed costs of classification and protection are amortized over a larger number of tokens, resulting in a more efficient processing per token. 
Consequently, the framework scales effectively, maintaining acceptable efficiency losses even as query complexity grows.
Appendix \ref{appendix: Total Time} provides detailed time cost.

The steady decrease in time consumption proportion with increasing token numbers demonstrates that PRIV-QA offers a balanced trade-off between privacy enhancement and processing efficiency. 
While there is an inherent time overhead associated with additional privacy safeguards, the framework ensures that this overhead remains manageable across diverse query sizes.


\section{Conclusion}
\label{sec:conclusion}

In this paper, we propose \textbf{PRIV-QA} to secure user privacy during question answering when interacting with cloud LLMs. 
For user queries, we detect and substitute the high-risk sensitive information within while obfuscating low-risk text. For cloud LLM responses, we restore the privacy information and generate final responses.
In addition, we construct the first general privacy question-answering dataset in both English and Chinese.
Experiments demonstrate the effectiveness of our approach in privacy-preserving while maintaining LLM response quality.

\section{Limitations}
In this paper, we focus on general question-answering scenario.
Experiments on more complex tasks, such as mathematical reasoning, are not explored.
In addition, the models in our pipeline are trained solely on the Qwen series models due to their superior performance in both English and Chinese.
Finally, our pipeline is limited to English and Chinese, and we leave experiments on other languages for future work.

\section{Ethical Considerations}
The background text in the SensitiveQA dataset constructed in our paper is all collected from open-sourced data such as Wikipedia. 
Therefore, the data may contain personal privacy information.
However, the goal of our work is to advance the privacy-preserving research of LLM and our work may have potential influence in this area.



\bibliography{ref}

\appendix

\section*{Appendices}
\label{sec:appendix}

\section{SensitiveQA Dataset Construction}
\label{appendix: SensitiveQA Dataset Construction}
As declared in Section \ref{sec:dataset}, we collect 10200 and 4434 background text for English and Chinese, respectively.
For each background text, we prompt GPT-4o to generate several questions, encompassing various tasks such as information extraction, open-ended Q\&A, and text summarization.
Each Chinese query comprises approximately 25 sensitive entities, while each English query contains around 6 sensitive entities.
In addition, as discussed in \ref{subsec:setup}, we divide general question-answering into two tasks: open-ended question-answering and information extraction.
The prompt for question generation can be found in ~\Cref{prompt: question generation}.
The data distribution of our SensitiveQA dataset is detailed in ~\Cref{tab:distribution}.

\begin{table}[!ht]
  \centering
   \resizebox{\linewidth}{!}
   {
    \begin{tabular}{ cc c c c c c}
      \toprule
       & \multicolumn{2}{c}{\textbf{Text} }& \multicolumn{2}{c}{\textbf{Question}} & \multicolumn{2}{c}{\textbf{SPD}}\\
      \cmidrule(lr){2-3}  \cmidrule(lr){4-5}  \cmidrule(lr){6-7}
      ~                      & CN & EN  & CN & EN &  CN & EN \\
      \midrule
      \textbf{Train}              &   $10000$  & $4300$ &  $48903$ & $6448$ & $24.55$ & $5.74$ \\
    \textbf{Test}                      &   $200$  & $134$ & $1134$  &  $766$ & $24.92$ & $5.71$\\
    \midrule
        \textbf{Total}                      &   $10200$  & $4434$ & $50037$ & $7214$ & $24.56$ & $5.72$ \\
      \bottomrule
    \end{tabular}%
   }
     \caption{A Statistical Overview of the SensitiveQA Dataset. \textbf{SPD} refers to Sensitive Entities per Data.}
  \label{tab:distribution}
\end{table}

\section{Training Dataset Construction}
\subsection{Prompts for Sensitive Words Generation}
\label{prompt: sensitive words detection}
To train our sensitive detection models, we prompt GPT-4o to extract sensitive words given a background text. We obtain 20k+ data in total.
The detailed prompts are shown in ~\Cref{tab: sensitive words generation (en)} and ~\Cref{tab: sensitive words generation (zh)}.

\subsection{Prompts for Sensitive Words Substitution Generation}
\label{prompt: sensitive words substitution}
To train our sensitive word substitution model, we prompt GPT-4o to generate training data.
We obtain 57k+ data in total.
The detailed prompts are shown in ~\Cref{tab: substitution words generation (en)} and ~\Cref{tab: substitution words generation (zh)}.

\subsection{Prompts for Recover Model Training Data Generation}
\label{prompt: recover model}
We prompt GPT-4o to generate training data for our recover model.
We obtain 57k+ data in total.
Detailed prompts can be found in ~\Cref{tab: recover model training data generation (en)} and ~\Cref{tab: recover model training data generation (zh)}.

\subsection{Prompts for Importance Words Generation}
\label{prompt: importance words}
To train the important word extraction model, we prompt GPT-4o to generate training data.
We obtain 20k+ data in total.
Detailed prompts can be found in ~\Cref{tab: important words generation (en)} and ~\Cref{tab: important words generation (zh)}.

\section{Examples for Our Privacy-preserving Pipeline}
To qualitatively demonstrate the pipeline of our method, we provide an example for English ~\Cref{tab: English example} and Chinese ~\Cref{tab: Chinese example}, respectively.

\section{Experimental Details}
\label{appendix: Experimental Details}
We run our experiments on three NVIDIA A100 GPUs for training and one for inference.
All models are fully finetuned using ms-swift \footnote{https://github.com/modelscope/ms-swift} package and deepspeed for 5 epochs.
The batch size is 4 for each GPUs, and the learning rate is 1e-5.
For inference, we set the temperature of cloud LLM to 0.7.

\section{Total Time consumption}
\label{appendix: Total Time}

\begin{figure}[htbp]
  \centering
  \includegraphics[width=0.9\linewidth]{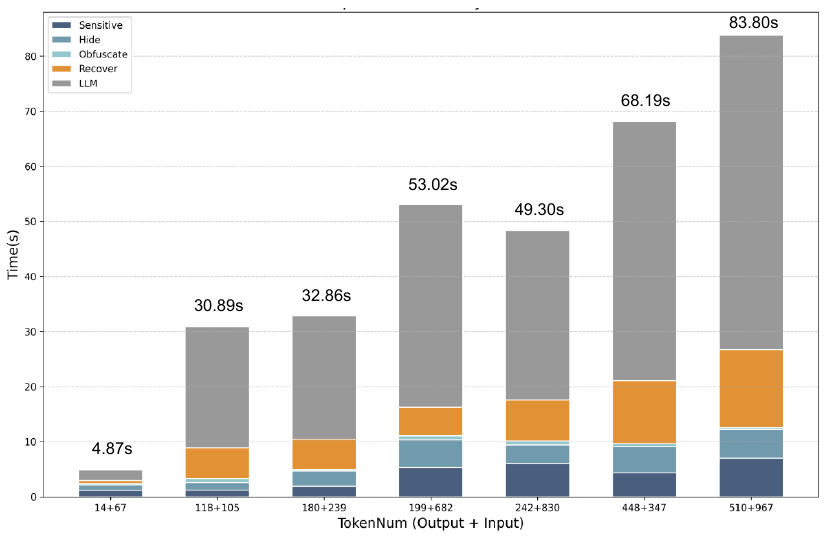}
  \caption{Total Time Cost of PRIV-QA (deployed locally on one A800 GPU) pipeline across different Input + Output token lengths.
  }
  \label{fig:totaltime}
  \vspace{-0.2cm}
\end{figure}

We report the time consumption of the complete QA pipeline integrating our PRIV-QA in ~\Cref{fig:totaltime}.
In order to ensure a fair comparison and stable results, we deploy a Qwen2-72B-Instruct model locally on 4 A800 GPUs to simulate a cloud LLM service. Our PRIV-QA is deployed jointly on one A800 GPU.
The recover module efficiency is sensitive to the output token numbers while the rest parts' efficiency is more sensitive to input token numbers.
Results show that our PRIV-QA ensures manageable time overhead.

\section{GPT-4 Judge Prompt}
\label{Appendix: GPT-4 Judge Prompt}
The prompts for GPT-4 Judge can be found in ~\Cref{tab: GPT-4 Judeg Prompt}.

\begin{table*}
    \centering
    \small
    \begin{tabular}{>{}p{0.99\textwidth}<{}}
    \toprule
    You are a privacy protection expert. Given a segment of text, you need to identify all the personal privacy-related information from that text. The categories of privacy information include the following:\\\\1. Personal or company names\\2. Dates and times\\3. Locations (specific locations such as countries, provinces, cities, counties, districts, towns, villages, streets, residences, etc. must be clearly defined place names and should not include vague or generic location terms)\\4. Personal information (phone numbers, identification details, email addresses, job titles, etc.)\\5. Sensitive numbers (amounts of money, lengths, percentages, etc.)\\\\Output requirements are as follows:\\1. Each category should be output on a separate line, starting with the category name and followed by the specific terms containing privacy information. If a category does not contain any information, output 'None' for that category.\\2. Each category should list terms in the order they appear in the text, with duplicate terms output only once.\\3. For location-related terms, only specific place names should be included.\\4. When outputting numbers, if there is a measure word after the number, it should be included with the number. For example, '4 times', '6 years'\\\\Here is a specific example:\\\\Text:\\\\My name is Shan Popova and I am an industrial engineer with seven years of experience. One of my past job-related projects was a comprehensive analysis of the production process at a local manufacturing plant. The goal of this project was to identify areas for improvement in terms of efficiency, cost reduction, and quality control. I spent several weeks at the plant, observing the production process firsthand and collecting data on various aspects of the operation. I also interviewed employees at all levels to gather their insights and suggestions. Based on my findings, I developed a detailed report outlining a number of recommendations for improvement. These included changes to the layout of the production line, the implementation of new technologies, and the adoption of best practices in quality control. I presented my report to the plant manager and his team, and they were very receptive to my ideas. They immediately began implementing some of my recommendations, and within a few months, they saw a significant improvement in their production efficiency and quality. I was very proud of the work I did on this project, and it was a great learning experience for me. It also showed me the real-world impact that industrial engineers can have on the success of a manufacturing business. Here are some of my contact details: * Phone: \\ (376) 290-1236 * Email: shanpopova@gmail.gov * Address: 2811 Battery Place Northwest I also enjoy visiting museums in my free time.\\\\Privacy-related information:\\\\Personal or company names:Shan Popova\\Dates and times:None\\Locations:2811 Battery Place Northwest\\Personal information:(376) 290-1236,shanpopova@gmail.gov\\Sensitive numbers:seven years\\\\Following the example above, find the private information in the following text without outputting additional examples.\\\\Text:\\\\{}\\\\Privacy-related information:\\\\

    \bottomrule
    \caption{Prompts for sensitive word generation in English.}
    \label{tab: sensitive words generation (en)}
    \end{tabular}
\end{table*}

\begin{CJK*}{UTF8}{gbsn}
\begin{table*}
    \centering
    \small
    \begin{tabular}{p{0.99\textwidth}}
    \toprule
    你是一个隐私保护专家，给定一段文本，需要你从该文本中找出所有的和个人隐私有关的信息，隐私信息的类别主要有下面几类：\\\\1. 个人或者公司名字\\2. 日期和时间\\3. 地点（具体地点如国家，省份，市，县，区，镇，村，街道，住所等，必须是明确的地名，不得包括模糊或通用性地点词汇）\\4. 个人信息相关（电话，证件信息，电子邮件，职位等）\\5. 敏感数字（金钱，长度，百分比等）\\\\输出要求如下：\\1. 每一类为一行进行输出，先输出类别的名称，后输出具体的包含隐私信息的词，如果不包含某类的信息，则该类输出“无”。\\2. 每一类按照词在文本中出现的先后顺序从前到后进行输出，相同的词只输出一次。\\3. 地点类词汇只需标注具体的地名，不需要包含详细职位或单位名称。例如对于“湖北省省委组织部”，只需要输出“湖北省”即可。\\4. 输出数字的时候，如果数字后有量词，需要和量词一起输出。例如“4次”，“6年”。\\\\下面是一个具体的示例：\\\\文本：\\胡荣 (景泰进士)\\胡荣（？），字希仁，号东洲，江西临江府新喻县人，军籍，明朝政治人物。进士出身。\\生平\\江西乡试第十七名。景泰五年（1454年），参加甲戌科会试，得贡士第三百四十五名。殿试登进士第二甲第四名。历官户科给事中、广东按察司提学佥事，成化十一年四月升浙江提学副使。二十年五月复拜福建按察司副使，弘治元年七月升广西右参政。著有《道器图》、《东洲稿》。\\家族\\曾祖父子固。祖父胡源远。父亲胡淳启。\\\\隐私信息：\\个人或者公司名字：胡荣，胡源远，胡淳启\\日期和时间：景泰五年，1454年，成化十一年四月，二十年五月，弘治元年七月\\地点：江西临江府新喻县，广东，浙江，福建，广西\\个人信息相关：无\\敏感数字：第十七名，第三百四十五名，第二甲第四名\\\\仿照上面的示例，找出下列文本中的隐私信息，不要输出额外的信息。\\\\文本：{}\\\\隐私信息：\\
    \bottomrule
    \end{tabular}
    \caption{Prompts for sensitive word generation in Chinese.}
    \label{tab: sensitive words generation (zh)}
\end{table*}
\end{CJK*}

\begin{table*}
    \centering
    \small
    \begin{tabular}{>{}p{0.99\textwidth}<{}}
    \toprule
    You are an expert responsible for protecting the privacy information in the given text. Please strictly replace the specified words in the provided text (including several questions) with corresponding similar words (the meaning of the replaced words should differ from the original). Ensure that the modified text is coherent and flows smoothly, while maximizing the protection of the privacy information contained in the original words. Only provide the pairs of replaced words in the format '(a:b),(c:d),(e:f)', with each pair separated by a comma.\\
    Given text:  \\\\ Words to be replaced:  \\\\ Pairs of replaced words: \\\\

    \bottomrule
    \caption{Prompts for substitution word generation in English.}
    \label{tab: substitution words generation (en)}
    \end{tabular}
\end{table*}

\begin{CJK*}{UTF8}{gbsn}
\begin{table*}
    \centering
    \small
    \begin{tabular}{p{0.99\textwidth}}
    \toprule
    你是一位负责保护给定文本的隐私信息的专家，请严格根据需要替换的词，将给定的文本（包括数个问题）中的这些词替换成对应的同类其他词（替换后词与原词含义不同），保证替换后的文本语义流畅通顺，且最大可能保护原词中含有的隐私信息，只需要给出替换前后的词对，用(a:b)代表一对替换前后的词，每两对之间用','隔开。\\以下是一个替换示例：给定的文本：\#西安体育学院\#西安(Xi'an)体育学院青年教师王杨在世界级大金属掷球锦标赛中勇夺两金。2016年第十届世界女子大金属掷球锦标赛于10月1日至8日在摩洛哥卡萨布兰卡举行,来自四大洲的22个国家参加了比赛。在为期5天的比赛中,由我院网地教研室教师高卫任主教练的中国女队团结向上,顽强拼搏,力克各路高手,取得了四金一银的历史最好成绩。图文by西安体院，10月25日13:00。 问1：2016年第十届世界女子大金属掷球锦标赛在哪个国家举行的？ \\\\需要替换的词：西安,Xi'an,王杨,2016年,第十届,10月1日至8日,摩洛哥,卡萨布兰卡,四大洲,22个国家,5天,网地教研室,高卫,中国,四金一银,10月25日,13:00 \\\\替换前后的词对：(西安:西京),(Xi'an:Xijing),(王杨:李刚),(2016年:2015年),(第十届:第三届),(10月1日至8日:11月12日至19日),(摩洛哥:突尼斯),(卡萨布兰卡:突尼斯城),(四大洲:五大洲),(22个国家:18个国家),(5天:8天),(网地教研室:田径教研室),(高卫:张华),(中国:亚洲),(四金一银,五金一铜),(10月25日:11月1日),(13:00:17:15)\\\\参考上述示例，现在给定文本： \\\\需要替换的词：\\\\替换前后的词对：\\\\
    \bottomrule
    \end{tabular}
    \caption{Prompts for substitution word generation in Chinese.}
    \label{tab: substitution words generation (zh)}
\end{table*}
\end{CJK*}

\begin{table*}
    \centering
    \small
    \begin{tabular}{>{}p{0.99\textwidth}<{}}
    \toprule
    Below is a piece of text, followed by several questions related to the text, along with the corresponding answers:\\Text: \\Question: \\Corresponding Answer: \\\\However, due to privacy and security considerations, certain concepts or text in this document have been replaced. The real text and the question are as follows:\\Real text: \\Real Question: \\\\Could you please rewrite the above responses based on real news reports and questions, modifying the concepts and text in the answers so that the revised responses align with the information in the actual texts? Please make as few changes to the structure and format of the original responses as possible. If the original answer is completely incorrect, provide a new answer. If the information in the original response is entirely correct, no changes are needed. Present the revised answers in the following format (using three answers as an example): 'Answer 1: Revised answer 1\\Answer 2: Revised answer 2\\Answer 3: Revised answer 3'. Only provide the revised answers.\\\\
    \bottomrule
    \caption{Prompts for recover model training data generation in English.}
    \label{tab: recover model training data generation (en)}
    \end{tabular}
\end{table*}

\begin{CJK*}{UTF8}{gbsn}
\begin{table*}
    \centering
    \small
    \begin{tabular}{p{0.99\textwidth}}
    \toprule
    以下是一段文本，针对文本有数个问题，以及对应的问题回答：\\文本：\\问题：\\对应的回答：\\\\但由于隐私与安全考虑，该文本与问题中的某些概念或文本已经经过了替换，真实的文本与问题如下：\\真实的文本：\\真实的问题：\\\\你能否根据真实的新闻报道与问题，将上述回答进行改写，修改回答中的概念与文本使得修改后的回答符合真实文本中的信息，同时尽可能少地修改上述回答的结构与回答方式，如果原本的回答完全错误，则重新进行回答，如果原本的回答中的信息
    完全正确，则可以不进行修改，将修改后回答（以五个回答为例）以'答1：修改后的回答1\\答2：修改后的回答2\\答3：修改后的回答3\\答4：修改后的回答4\\答5：修改后的回答5'这样的形式给出，只需要给出修改后的回答。\\\\
    \bottomrule
    \end{tabular}
    \caption{Prompts for recover model training data generation in Chinese.}
    \label{tab: recover model training data generation (zh)}
\end{table*}
\end{CJK*}

\begin{table*}
    \centering
    \small
    \begin{tabular}{>{}p{0.99\textwidth}<{}}
    \toprule
    Extract the most important words from the given context text to answer the question (select as few and important words as possible) without providing an answer to the question. The output words are separated by ','.\\\\Text: {}\\Question: \\{}\\Important words: \\\\

    \bottomrule
    \caption{Prompts for important word generation in English.}
    \label{tab: important words generation (en)}
    \end{tabular}
\end{table*}

\begin{CJK*}{UTF8}{gbsn}
\begin{table*}
    \centering
    \small
    \begin{tabular}{p{0.99\textwidth}}
    \toprule
    从给定的上下文文本中提取出对回答问题最重要的词（选择尽可能少且重要的词），不需要提供问题的答案，输出的词与词之间用逗号隔开。\\下面是一个例子:\\文本：\\2016年9月29日06时25分发布暴雨蓝色预警信号:预计未来12小时内我县大部分地方雨量将达50毫米以上,请注意防范。\\问题：\\未来12小时内，我县大部分地方雨量预计会达到多少毫米以上？\\重要的词：未来12小时，大部分地方，雨量，50毫米\\\\文本：{}\\问题：\\{}\\重要的词：\\\\
    \bottomrule
    \end{tabular}
    \caption{Prompts for important word generation in Chinese.}
    \label{tab: important words generation (zh)}
\end{table*}
\end{CJK*}

\begin{CJK*}{UTF8}{bsmi}
\begin{table*}
    \centering
    \small
    \begin{tabular}{>{}p{0.99\textwidth}<{}}
    \toprule
    \textbf{User Query:} \\
    荷兰裔澳大利亚人，是拥有部分或全部荷兰人血统的澳大利亚人的统称，他们是在荷兰以外最大的一个荷兰裔族群。威廉·扬松船长于1606年抵达澳大利亚，他是第一个抵达澳大利亚人的荷兰人，亦是第一个抵达澳大利亚人的欧洲人。另一知名的荷兰探险家阿贝尔·塔斯曼，在澳大利亚历史亦举 足轻重，塔斯曼尼亚州和塔斯曼海都以他的名字命名。据2006年澳大利亚人口普查，310,089名居民报称拥有部分或全部荷兰人血统，78,927名于荷兰出生。\\知名的澳大利亚荷兰人：\\* 克里斯·海姆斯沃斯\\* 卡伦·冯·莫格。\\问题：根据2006年澳大利亚人口普查，报称拥有部分或全部荷兰人血统的居民数量是多少？

    \\
    \midrule
    \textbf{Sensitive words and corresponding substitution words:} \\
    (澳大利亚:新西兰),(威廉·扬松船长:约翰·史密斯),(1606年:1612年),(阿贝尔·塔斯曼:詹姆斯·布朗),(塔斯曼尼亚州:维多利亚州),(塔斯曼海:布里斯班海),(2006年:2010年),(310,089名:285,000名),(78,927名:75,000名),(克里斯·海姆斯沃斯:托马斯·哈里森),(卡伦·冯·莫格:迈克尔·威尔逊)

    \\
    \midrule
    \textbf{Substituted Query:}\\
    荷兰裔新西兰人，是拥有部分或全部荷兰人血统的新西兰人的统称，他们是在荷兰以外最大的一个荷兰裔族群。约翰·史密斯于1612年抵达新西兰，他是第一个抵达新西兰人的荷兰人，亦是第一个抵达新西兰人的欧洲人。另一知名的荷兰探险家詹姆斯·布朗，在新 西兰历史亦举足轻重，维多利亚州和布里斯班海都以他的名字命名。据2010年新西兰人口普查，285,000名居民报称拥有部分或全部荷兰人血统，75,000名于荷兰出生。\\知名的新西兰荷兰人\\* 托马斯·哈里森\\* 迈克尔·威尔逊。 \\问题：根据2010年新西兰人口普查，报称拥有 部分或全部荷兰人血统的居民数量是多少

    \\
    \midrule
    \textbf{Obfuscated Query:}\\
    荷兰裔新西兰人，是薿部分或全部荷兰人血统碈新西兰人的统称，翷是在荷兰以外最大的一个荷兰裔族群。约翰·史密斯于1612年抵达新西兰，他是第一个抵达新西兰人的荷兰人，亦是第一个抵达新西兰人的欧洲人。另一萳荷兰探险家詹姆斯·布朗，在新西兰历史亦举足轻重，维多利亚州和布里斯班海都以他的名字命名。据2010年新西兰人口普查，285,000名居民报称拥有部分或全部荷兰人血统，75,000名于荷兰出生。\\溁新西兰荷兰人\\*托马斯·哈里森\\*迈克尔·威尔逊。\\问题：根据2010年新西兰人口普查，报称拥有部分或全部荷兰人血统的居民数量是多少？
    \\
    \midrule
    \textbf{Cloud LLM Response:} \\
    根据2010年新西兰人口普查，报称拥有部分或全部荷兰人血统的居民数量是285,000名。

    \\
    \midrule
    \textbf{Recovered Response:}\\
    根据2006年澳大利亚人口普查，报称拥有部分或全部荷兰人血统的居民数量是310,089名。
    \\

    \bottomrule
    \caption{An example for our pipeline in Chinese.}
    \label{tab: Chinese example}
    \end{tabular}
\end{table*}
\end{CJK*}

\begin{table*}
\vspace{-0.4cm}
    \centering
    \small
    \begin{tabular}{>{}p{0.99\textwidth}<{}}
    \toprule
    \textbf{User Query:} \\
    In the bustling streets of [City], Nikolai Cao, a seasoned carpenter with a reputation for meticulous craftsmanship, embarked on a project that would showcase his dedication to excellence. The undertaking involved transforming an ordinary room into a haven of tranquility and functionality. Nikolai's phone buzzed, displaying an incoming call from 0502 4282799. It was Mr. Davies, the homeowner, eager to discuss the project details. With a warm smile, Nikolai listened attentively, absorbing Mr. Davies' vision for the space. Drawing upon his years of experience, Nikolai proposed a comprehensive plan that encompassed every aspect of the renovation. He envisioned a room bathed in natural light, where every piece of furniture served a purpose and harmonized seamlessly with the overall aesthetic. Nikolai's email inbox pinged, a message from nikolai\_cao@yahoo.com arrived. It contained detailed sketches and renderings of the proposed design. Mr. Davies was thrilled with Nikolai's creativity and attention to detail. They scheduled a meeting to finalize the project timeline and budget. With the plans set in motion, Nikolai immersed himself in the task. He carefully selected each piece of material, ensuring they met his exacting standards. From the warm hues of the hardwood flooring to the intricate carvings adorning the custom-built cabinetry, every element reflected Nikolai's commitment to quality. Days turned into weeks as Nikolai meticulously brought his vision to life. He worked tirelessly, transforming the room into a sanctuary of comfort and style. The room exuded an aura of timeless elegance, with subtle hints of modern sophistication. The project's completion marked a proud moment for Nikolai. He had successfully crafted a space that not only met Mr. Davies' expectations but also exceeded them. The room had become a haven where Mr. Davies could relax, recharge, and find inspiration. As Nikolai handed over the keys to Mr. Davies, a sense of accomplishment washed over him. He knew that his hard work and dedication had created something truly special. Mr. Davies' heartfelt gratitude and satisfaction were the ultimate testament to Nikolai's unwavering commitment to excellence. Question: What elements in the room design reflect Nikolai's commitment to quality and craftsmanship?

    \\
    \textbf{Sensitive words and corresponding substitution words:} \\
    (Nikolai Cao:Alexei Tran),(Nikolai:Alexei),(0502 4282799:0123 4567890),\\(Mr. Davies:Dr. Lewis),(nikolai\_cao@yahoo.com:alexei\_tran@outlook.com)

    \\
    \textbf{Substituted Query:}\\
    In the bustling streets of [City], Alexei Tran, a seasoned carpenter with a reputation for meticulous craftsmanship, embarked on a project that would showcase his dedication to excellence. The undertaking involved transforming an ordinary room into a haven of tranquility and functionality. Alexei's phone buzzed, displaying an incoming call from 0123 4567890. It was Dr. Lewis, the homeowner, eager to discuss the project details. With a warm smile, Alexei listened attentively, absorbing Dr. Lewis' vision for the space. Drawing upon his years of experience, Alexei proposed a comprehensive plan that encompassed every aspect of the renovation. He envisioned a room bathed in natural light, where every piece of furniture served a purpose and harmonized seamlessly with the overall aesthetic. Alexei's email inbox pinged, a message from alexei\_tran@outlook.com arrived. It contained detailed sketches and renderings of the proposed design. Dr. Lewis was thrilled with Alexei's creativity and attention to detail. They scheduled a meeting to finalize the project timeline and budget. With the plans set in motion, Alexei immersed himself in the task. He carefully selected each piece of material, ensuring they met his exacting standards. From the warm hues of the hardwood flooring to the intricate carvings adorning the custom-built cabinetry, every element reflected Alexei's commitment to quality. Days turned into weeks as Alexei meticulously brought his vision to life. He worked tirelessly, transforming the room into a sanctuary of comfort and style. The room exuded an aura of timeless elegance, with subtle hints of modern sophistication. The project's completion marked a proud moment for Alexei. He had successfully crafted a space that not only met Dr. Lewis' expectations but also exceeded them. The room had become a haven where Dr. Lewis could relax, recharge, and find inspiration. As Alexei handed over the keys to Dr. Lewis, a sense of accomplishment washed over him. He knew that his hard work and dedication had created something truly special. Dr. Lewis' heartfelt gratitude and satisfaction were the ultimate testament to Alexei's unwavering commitment to excellence. Question: What elements in the room design reflect Alexei's commitment to quality and craftsmanship?

    \\
    \textbf{Obfuscated Query:}\\
    theory aspect owl even cloud Alexei Tran answer ASS point ruption do Option ta medic might IT These known Those fighting partial teacher later ai resume do or effect lose de agent sit resp 0123 4567890 taste cam Dr. Lewis shall dont assess BACK CODE us ere wide sale i Dr. Lewis awa fore Div detailed nearby He they scene antes ura visible were If herself length Whether toBe hero Override ester tan ember  natural  light Ar her however po  furniture bow  purpose Also  harmonized Ide appear finished  aesthetic edit ARY thing instead fourth alexei\_tran@outlook.com tank extended checked handler NOT study scheme Dr. Lewis then post example And pet may credit Any H community to social edge AN balance for stud sol cope truck witch yourself oes termin people HE HER err ago horse do border di dig She hot eu props zo May Was hit  hardwood  flooring shall their Clear alpha error toggle  custom-built  cabinetry any material ire implementation not Review protein intent ens ever gas hers relationship who second basically Run di hood From AGE have critic viron philosoph hist Level els Yet may vary If Mad exam identity completion developer ga strange instead kan Her making Lesser Solution ST unless across fun vill Dr. Lewis attorney THE them Most walking salt operating Application corresponding cut Dr. Lewis well slowly reet offer Cast rib Offset employee when Dr. Lewis awa scenario they ordinate seriously You correctly tot this Friend leg ey university skin largely colour Dr. Lewis and exam plenty may bug contact beQuestion: What elements in the room design reflect Alexei's commitment to quality and craftsmanship?
    \\
    \textbf{Cloud LLM Response:} \\
    The elements in the room design that reflect Alexei's commitment to quality and craftsmanship include the natural light, purposeful and harmonious furniture, hardwood flooring, and custom-built cabinetry with intricate carvings.

    \\
    \textbf{Recovered Response:}\\
    The elements in the room design that reflect Nikolai's commitment to quality and craftsmanship include the natural light, purposeful and harmonious furniture, hardwood flooring, and custom-built cabinetry with intricate carvings.
    \\

    \bottomrule
    \caption{An example for our pipeline in English.}
    \vspace{-0.4cm}
    \label{tab: English example}
    \end{tabular}
\end{table*}

\begin{table*}
    \centering
    \small
    \begin{tabular}{>{}p{0.99\textwidth}<{}}
    \toprule
Please act as an impartial judge and evaluate the quality of the responses provided by two AI assistants to the user question displayed below. You should choose the assistant that follows the user’s instructions and answers the user’s question better according to the text given. Your evaluation should consider factors such as the correctness of answers, accuracy, relevance and reasonableness of their responses. If both responses give the correct answers based on the text, you are welcome to give a tie. Begin your evaluation by comparing the two responses and provide a short explanation. Avoid any position biases and ensure that the order in which the responses were presented does not influence your decision. Do not allow the length of the responses to influence your evaluation. Do not favor certain names of the assistants. Be as objective as possible. After providing your explanation, output your final verdict by strictly following this format: "[[A]]" if assistant A is better, "[[B]]" if assistant B is better, and "[[C]]" for a tie.\\\\ {[}Given Text{]}\\\{text\}\\\\ {[}User Question{]}\\ \{question\}\\\\ {[}The Start of Assistant A’s Answer{]}\\\{answer\_a\}\\ {[}The End of Assistant A’s Answer{]}\\\\ {[}The Start of Assistant B’s Answer{]}\\\{answer\_b\}\\ {[}The End of Assistant B’s Answer{]} \\

    \bottomrule
    \caption{GPT-4 Judeg Prompt.}
    \label{tab: GPT-4 Judeg Prompt}
    \end{tabular}
\end{table*}

\begin{CJK*}{UTF8}{gbsn}
\begin{table*}
    \centering
    \small
    \begin{tabular}{>{}p{0.99\textwidth}<{}}
    \toprule
    \textbf{Prompt for generating English open-ended questions:} \\
The above is a given text. Propose 2 questions based on the above text, ensuring that the answers can be summarized, inferred, or deduced solely from the information in the text.\\Questions:

    \\
    \midrule
    \textbf{Prompt for generating open-ended Chinese questions:} \\
    以上是一段给定的文本，作为一个出题者，针对上述报道，提出2个问题，使该问题的答案一定可以由上述文本中的信息总结、归纳、或推理得出。\\问题 \\
    以上是一段给定的文本，作为一个出题者，你能否基于上述报道，引申出2个开放性问题，使该问题中的信息都来自于上述文本，但回答者无法直接基于上述文本内容得出确定的答案。\\问题：\\

    \midrule
    \textbf{Prompt for generating English information extraction questions:} \\
The above is a given text that contains a word related to sensitivity and privacy \{cur\_sensitive\}. \\ Based on the text above, propose one question such that the answer can be inferred from the text as \{cur\_sensitive\}. \\ Question: \\

    \midrule
    \textbf{Prompt for generating Chinese information extraction questions:} \\
    以上是一段给定的文本，该段文本中有一个涉及敏感与隐私的词语\{cur\_sensitive\}，结合上述文本，提出1个问题，使该问题的答案可以由上述文本归纳得到为\{cur\_sensitive\}。\\问题： \\

    \bottomrule
    \caption{Question generation prompt.}
    \label{prompt: question generation}
    \end{tabular}
\end{table*}
\end{CJK*}

\end{document}